\documentclass{article}

\usepackage{arxiv}

\usepackage[utf8]{inputenc} 
\usepackage[T1]{fontenc}    
\usepackage{hyperref}       
\usepackage{url}            
\usepackage{booktabs}       
\usepackage{amsfonts}       
\usepackage{nicefrac}       
\usepackage{microtype}      
\usepackage{lipsum}		
\usepackage{graphicx}
\usepackage{natbib}
\usepackage{doi}

\usepackage{subcaption}
\usepackage{csquotes}
\usepackage{amsmath}
\usepackage{cleveref}

\title{Automating Wood Species Detection and Classification in Microscopic Images of Fibrous Materials with Deep Learning}

\date{} 					

\author{ \href{https://orcid.org/0000-0002-7523-5694}{\includegraphics[scale=0.06]{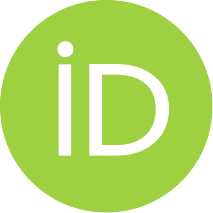}\hspace{1mm}Lars~Nieradzik}\\
	Image Processing Department\\
	Fraunhofer ITWM\\
	Fraunhofer Platz 1, 67663, Kaiserslautern\\
	\texttt{lars.nieradzik@itwm.fraunhofer.de}\\
	\And
	\href{https://orcid.org/0009-0001-7547-269X}{\includegraphics[scale=0.06]{orcid.pdf}\hspace{1mm}Jördis ~Sieburg-Rockel} \\
	Thünen Institute of Wood Research\\
	Leuschnerstraße 91, 21031, Hamburg\\
	\texttt{joerdis.sieburg-rockel@thuenen.de} \\
	\And
	\href{https://orcid.org/0009-0009-6611-3140}{\includegraphics[scale=0.06]{orcid.pdf}\hspace{1mm}Stephanie~Helmling} \\
	Thünen Institute of Wood Research\\
	Leuschnerstraße 91, 21031, Hamburg\\
	\texttt{stephanie.helmling@thuenen.de} \\
	\And
	\href{https://orcid.org/0000-0002-1327-1243}{\includegraphics[scale=0.06]{orcid.pdf}\hspace{1mm}Janis~Keuper} \\
	CC High Performance Computing\\
	Fraunhofer ITWM\\
	Fraunhofer Platz 1, 67663, Kaiserslautern\\
	\texttt{keuper@imla.ai}\\
	\And
	\href{https://orcid.org/0009-0009-5848-5024}{\includegraphics[scale=0.06]{orcid.pdf}\hspace{1mm}Thomas~Weibel} \\
	Image Processing Department\\
	Fraunhofer ITWM\\
	Fraunhofer Platz 1, 67663, Kaiserslautern\\
	\texttt{thomas.stephani@itwm.fraunhofer.de}\\
	\And
	\href{https://orcid.org/0009-0007-2249-2797}{\includegraphics[scale=0.06]{orcid.pdf}\hspace{1mm}Andrea~Olbrich} \\
	Thünen Institute of Wood Research\\
	Leuschnerstraße 91, 21031, Hamburg\\
	\texttt{andrea.olbrich@thuenen.de} \\
	\And
	\href{https://orcid.org/0000-0002-9821-1636}{\includegraphics[scale=0.06]{orcid.pdf}\hspace{1mm}Henrike~Stephani} \\
	Image Processing Department\\
	Fraunhofer ITWM\\
	Fraunhofer Platz 1, 67663, Kaiserslautern\\
	\texttt{henrike.stephani@itwm.fraunhofer.de}\\
}

\renewcommand{\shorttitle}{Automating Wood Species Detection and
Classification}

\hypersetup{
pdftitle={Automating Wood Species Detection and Classification in Microscopic Images of Fibrous Materials with Deep Learning},
pdfsubject={-},
pdfauthor={Lars Nieradzik, Jördis Sieburg-Rockel, Stephanie Helmling, Janis Keuper, Thomas Weibel, Andrea Olbrich, Henrike Stephani},
pdfkeywords={deep learning, wood identification, maceration, vessel elements, EU Timber Regulation},
}

\begin{document}
\maketitle

\begin{abstract}
	We have developed a methodology for the systematic generation of a large image dataset of macerated wood references, which we used to generate image data for nine hardwood genera. This is the basis for a substantial approach to automate, for the first time, the identification of hardwood species in microscopic images of fibrous materials by deep learning. Our methodology includes a flexible pipeline for easy annotation of vessel elements. We compare the performance of different neural network architectures and hyperparameters. Our proposed method performs similarly well to human experts. In the future, this will improve controls on global wood fiber product flows to protect forests.
\end{abstract}


\section{Introduction}
In order to reduce illegal logging, the European Union (EU) has, since 2013, required documentation on the origin and species of wood contained in every wood product that is placed on the EU market (EU Timber Regulation, EUTR, No. 995/2010). At the end of 2024, a new regulation is planned to be introduced to avoid global deforestation. The new EU regulation will target a wider range of products to ensure that these products are \enquote{deforestation-free} \citep{european2021proposal}.

To control the compliance with these laws, the demand for wood species identification is already high and expected to increase. 

Various methods have been developed to identify wood species, such as genetic analysis, near-infrared (NIR) spectroscopy, stable isotopes and wood anatomy \citep{schmitz2020overview}.
When it comes to identifying wood in non-solid samples, such as paper or pulp, microscopic analysis of the wood anatomy is the method of choice. 163 structural features have been defined by the International Association of Wood Anatomists \citep{wheeler1989iawa} and have been used for microscopic descriptions of about 8700 timbers collected in various databases \citep{Jorgo, insidewood, openagrar_mods_00085331}. Only a few of these structural features can still be utilized for the analysis of fibrous materials. Nevertheless, there are very good descriptions of the references available for the woods mainly used in paper production \citep{ilvessalo1995fiber, helmling2018atlas}.
While macroscopic wood analysis already requires extensive expert knowledge of wood anatomists \citep{ruffinatto2019atlas}, microscopic analysis requires more effort, as solid wood samples have to be prepared into thin cuts or the cells of a paper sample have to be individualized. The main challenge in the analysis of paper is that the structural features relating to the three-dimensional arrangement in the tissue cannot be detected from the individual cells of the fibrous materials. Therefore, the anatomical identification of woods is performed at most at the genus level. In addition, paper usually consists of a mixture of different genera. Therefore, two slides per sample must be systematically examined by wood anatomists in order to detect all wood genus present, even those in lower concentrations.

Due to the considerable amount of time required (and due to the limited number of competent scientists in this field), the experts are currently not able to fulfill the need for this analysis.
For this reason, the economic and ecological impact of improving and facilitating the analysis of these microscopic images is huge.

For more than 20 years, very good computer-aided wood species identification systems have been available, such as Commercial Timbers \citep{Jorgo}, Inside Wood \citep{insidewood} or CITESwoodID \citep{JorgoCITES}. Large databases, also available online. There have been endeavors in recent years to automatize macroscopic solid wood identification by using image based machine learning techniques. In particular, field-ready wood identification systems, such as MyWood-Premium \citep{MyWood}, XyloTron \citep{ravindran2020xylotron}, or XyloPhone \citep{wiedenhoeft2020xylophone} could contribute greatly to the fight against the illegal wood trade. This artificial intelligence (AI) is producing promising results and is currently under fast development. \citep{silva2022computer}. In microscopic wood identification especially for the analysis of fibrous material, on the other hand, the process is still mostly manual, and no automatic approach is yet known to the authors.\\

\begin{figure*}[t!]
\includegraphics[width=\textwidth]{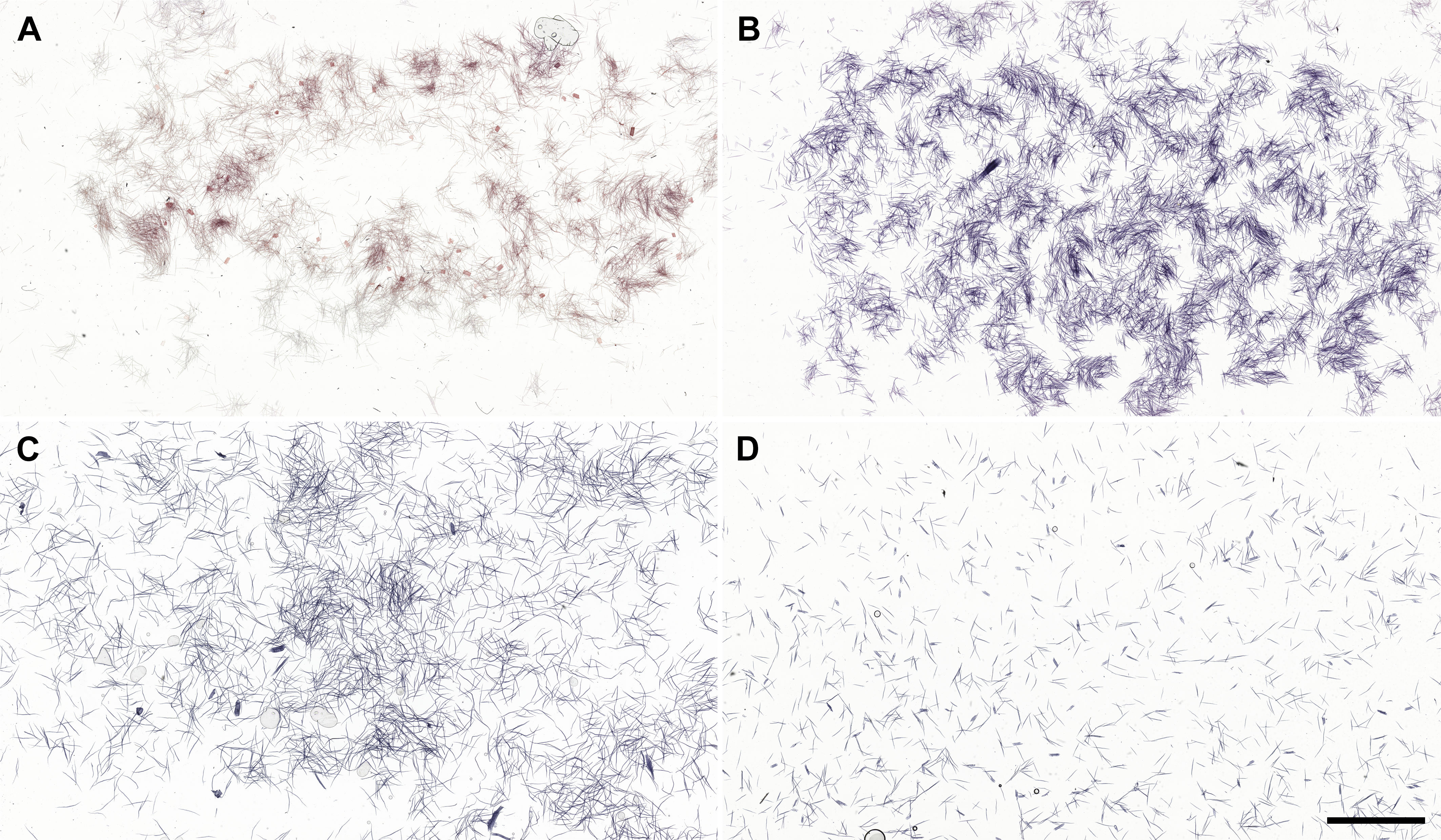}
\caption[Overview images]{Overview images of differently stained macerated samples of A \emph{Acacia}, B \emph{Populus}, C \emph{Hevea}, D \emph{Salix}. Scale bar=5mm.}
\label{fig:Overview}
\end{figure*}

In this paper, we present how a large number of reference specimens for hardwood fibers can be imaged across the entire slide in five focal planes to build the database. To facilitate the expert process of wood species identification, these data are used to train a deep-learning based system to analyze unknown samples automatically. With this automatization, more samples can be analyzed and protection of forests will be improved. \\
As any data-driven method relies on a solid, well annotated database, we present a methodology that enables easy annotation and reannotation to increase annotation throughput and minimizing this tedious work for biological experts. Furthermore, as we want to make use of the plethora of deep learning algorithms now and in the future, we use a pipeline with interchangeable single processing steps, that also have measurable performance. \\
We will show that our proposed method is able to solve the given task, as well as illustrate the influence, different networks and parametrizations have on the single processing steps. 
Hence, the key contributions of this paper are:
\begin{itemize}
    \item An initial, comprehensive microscopic image reference data-set for hardwood fiber material is available and can be easily extended using this method.
    \item We are the first to automate microscopic wood detection of hardwood fibers using neural networks. Our automated method performs similarly well to human experts.
    \item A flexible pipeline is presented on how the dataset was generated. This data is then used for training the networks. This pipeline can be used for additional hardwood samples/species in the future, and be extended to be used on mixed samples.
    \item We perform in-depth comparison of state-of-the-art neural network architectures and hyperparameters guided by biological domain expertise.
    \item With our tests, we also show how large microscopic data of this and similar types must be preprocessed to train neural networks. In particular, we show the effect of color channels, different focal planes, and image sizes on accuracy.
\end{itemize}

\section{Materials and Methods}

We first discuss the methods applied for sample preparation, the microscope that is used and the respective influences on image generation and variation.

We will then describe the process pipeline for automatic image analysis with state-of-the-art deep neural networks and its individual methods and parameters.

\subsection{Sample, Optics and Image Generation}

Sample preparation is very complex and requires biological and technical expertise. For reasons of feasibility, we focus on hardwood samples. They can be identified by vessel elements and their anatomical structure, as described in \cite{ilvessalo1995fiber} and \cite{helmling2018atlas}. Within those samples, we mainly selected commonly processed timbers that are cultivated in plantations for pulp and paper and fiber board production. To test the limitations of the method, morphologically very different as well as very similar species were selected. Vouchered specimens of the wood collection of the \emph{Thünen institute} and other documented sources served as reference material for training and testing. Analogously to pulp production, the cell compound of wooden tissue is dissolved into individual cells by maceration according to the method of \citep{franklin1945preparation}. Maceration and staining are described in \cite{helmling2016qualitative} and \cite{helmling2018atlas}. Alexander Herzberg solution and nigrosin (1 wt\%) were used for staining. As Alexander Herzberg staining is not durable, the slides must be examined without delay.\\
In a later real-life application scenario, the cell density on the slides will vary depending on the preparator. To consider this variance in the training data, the preparations were made by different people. 

\Cref{fig:Overview} shows four different overview images of macerated samples of different species. Density and color as well as shape of vessel elements show a high variance. Influencing factors are: choice of species, staining method, preparing  technician, preparation agents, density and microscope. To record the data and automatically digitize a large number of samples, we use the microscope slide scanner Axioscan 7 (Zeiss, Germany). With Objective N-Achroplan 5x/0,15 five focal levels per slide were recorded over an area of approximately 8 cm$^{2}$ with a scale per pixel of 0,69 x 0,69 x 16,33 µm$^{3}$ (Software, ZEN slidescan 3.5, Zeiss, Germany).
Depending on the material, the vessel elements and cell types can have different levels of destruction. They may still be completely intact, but may also be torn into smaller fragments.

\subsection{Overview of Algorithm Pipeline}

Convolutional neural networks (CNNs) are widely applied to all kinds of image processing problems for the last 15 years. They have proven to be well generalizable and generally outperform traditional computer vision methods \citep{o2019deep}. However, as in most real world applications, we are not presented with a fixed data set and a well-defined classification or segmentation task, but rather have to model and build the dataset ourselves. One of the key problems is that annotation of microscopic overview images of these samples is tedious. Additionally, it can only be performed by biological wood identification experts (wood anatomists) that are familiar with each species' characteristic cell structure. We annotated the vessel elements using ZEN blue 3.4 software from Zeiss, Germany.

\begin{figure}[t!]
\centering
\includegraphics[scale=0.17]{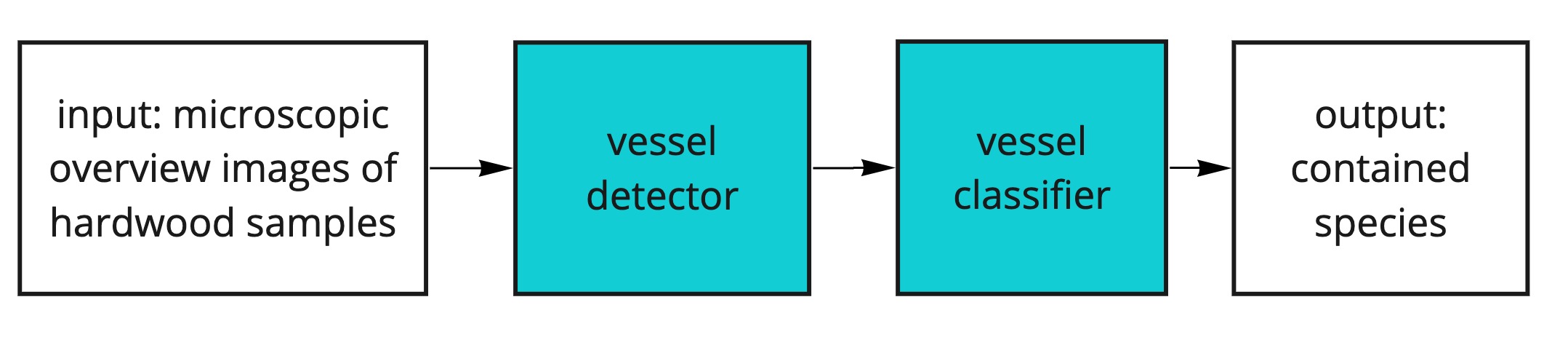}
\caption[Simple two-step procedure]{Two-step procedure where first vessel elements are detected and then classified}
\label{fig:TwoStep}
\end{figure}

All modern object detection networks such as Faster R-CNN \citep{DBLP:journals/corr/RenHG015}, DINO \citep{DBLP:journals/corr/abs-2104-14294}, DETR \citep{zhang2022dino} or YOLOv7 \citep{yolo} perform detection and classification in one step. We, however, use a two-step algorithm approach, as illustrated in \Cref{fig:TwoStep}. We first detect the vessel elements as bounding boxes and then perform classification on these bounding boxes as a second step. There are several reasons for doing this. The first reason is that it imitates the process that is done by visual-manual analysis. This improves interpretability and comparability with the previous human method. Secondly, we can assume that we do not require the same image resolution for detection of vessel elements as for their classification. Reducing the resolution, which can be up to $50 000^2$ pixels, is the easiest mean of reducing computational cost and increasing flexibility. Last but not least, it makes the generation of a database easier, as we explain in the next section.

\begin{figure}[t!]
\centering
\includegraphics[scale=0.07]{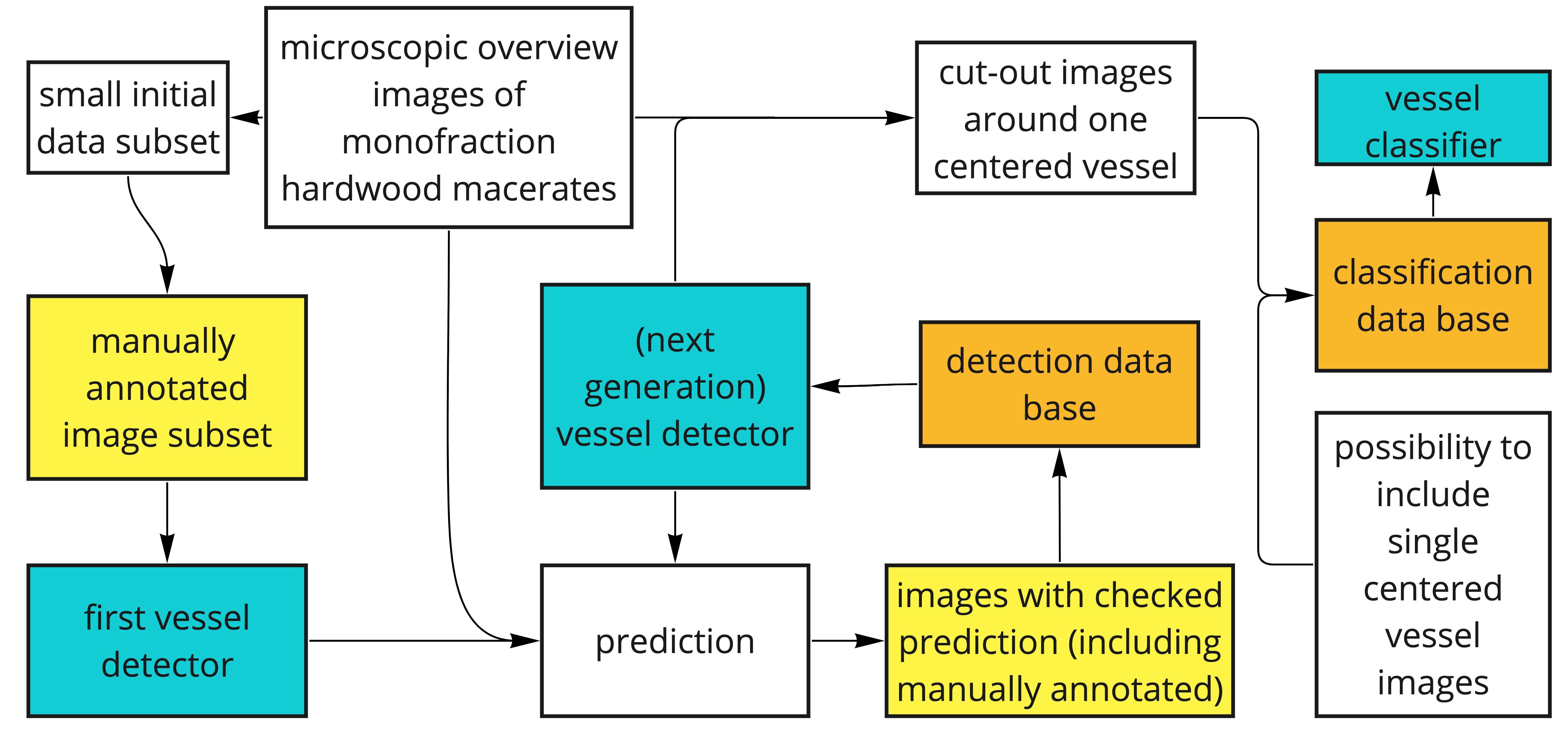}
\caption[Annotation]{Database building and learning pipeline. Wood expert annotation effort in yellow, databases are in orange and deep networks in blue}
\label{fig:Annotation}
\end{figure}

\subsubsection{Database Generation}
In order to develop a robust and generalizable deep learning procedure systematically, we want to build a large database quickly. A two-step procedure is very advantageous for that purpose. Illustrated in \cref{fig:Annotation}, one can see how the databases for the detector and classifier are trained iteratively to minimize effort for the annotators as much as possible: 
That is why we are starting to build up a database of overview images using only pure species samples, rather than directly using mixed species' materials and annotating them.

\begin{figure*}
    \centering
    \includegraphics[width=\textwidth]{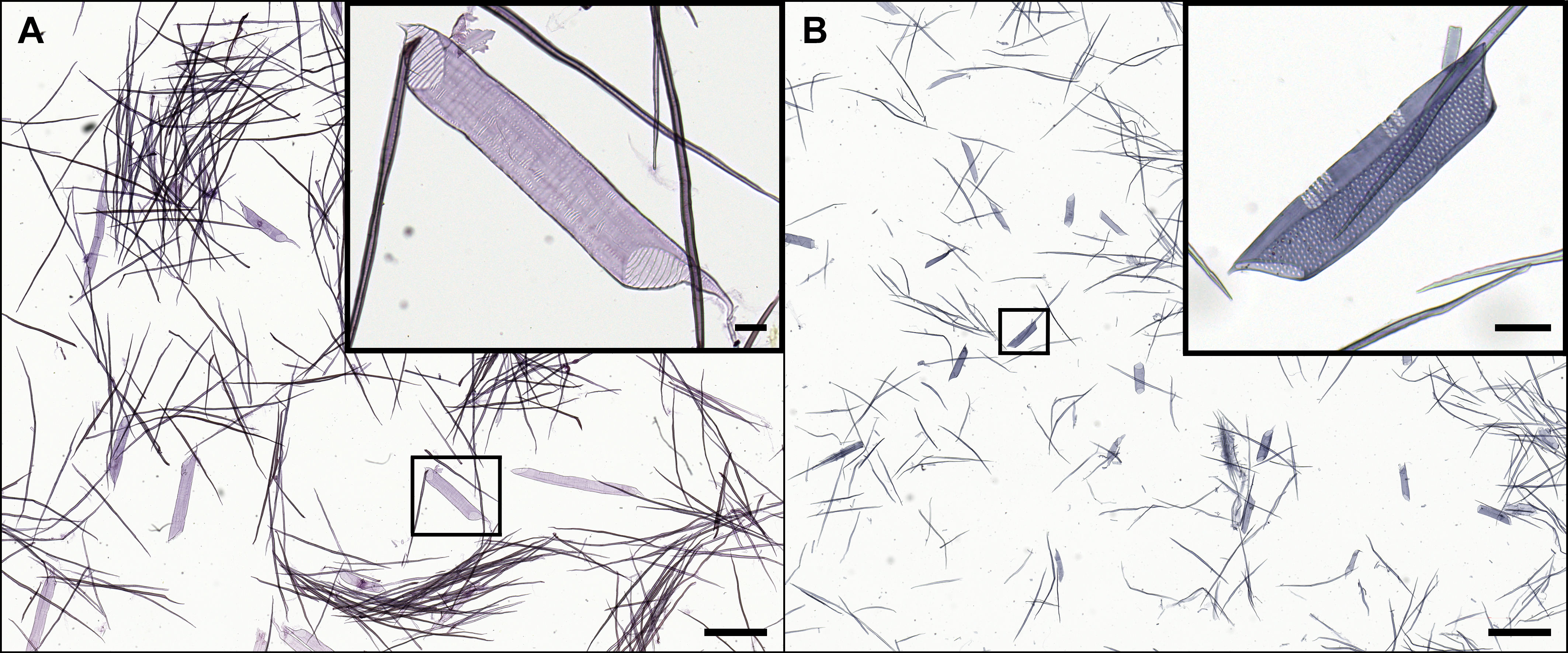}
    \caption{Image cutouts of vessel elements from overviews of A \emph{Schima} and B \emph{Populus}. Scale bars: overview=1mm, cutout=100µm.}
\label{fig:CutOut}
\end{figure*}

The advantage is two-fold: first of all, only location annotation must be performed, and classification annotation can be avoided. Secondly, pure samples are easier to come by than systematically mixed samples would be. 
Detection is then performed on the mono-fraction overview images, while classification is done on full resolution mostly centered cut-outs. The detection is trained with several genera to ensure the recognition of a greater habitus diversity of the vessel elements. These genera are \emph{Acacia} (Acacia), \emph{Betula} (Birch), \emph{Eucalyptus} (Eucalypt), \emph{Fagus} (Beech), \emph{Hevea} (Rubberwood), \emph{Liquidambar} (Sweet gum), \emph{Populus} (Poplar), \emph{Salix} (Willow) and \emph{Schima} (Chinese guger tree). \emph{Schima} and \emph{Populus} exemplarily shown in \cref{fig:CutOut}. However, one important aspect that has to be considered with special care is the so-called data leakage \citep{DBLP:journals/corr/abs-2102-11673}.

\subsubsection{Data leakage and dataset splitting}

When training a model, it is always necessary to split the data in a way that avoids training "wrong" features. Learning wrong features instead of relevant ones is called data leakage. One typical example of data leakage is the brightness of an image. If images of some classes tend to be darker than images of other classes, this can lead to the network only paying attention to the brightness of the image and not learning relevant features.

The maceration process as described above, produces samples that should have a variance that is characteristic for the respective species, we want to detect and classify. However, as maceration is a manual process, it can also include variance in the data that is characteristic for the individual human preparation instead of the species. Therefore, it is important not to learn these preparation differences. This is done by making sure that each genus is represented by at least three independent macerates. The respective images are then split to generate independent datasets for training, validation and testing.

The test dataset is used only for a final check, while the training and validation datasets are used for training and optimizing hyperparameters. We keep the same ratio of classes in both the training and validation splits (stratified).


\subsection{Detection}

The detection step consists of locating the vessel elements in the images. The final vessel element classification is based on fine structures within the vessel elements and therefore will only be possible with a high enough resolution.
At the same time, the detection of the vessel elements can be performed on downscaled images in a more efficient way. Hence, even if modern object detectors such as YOLOv7 \citep{yolo} or DINO \citep{zhang2022dino} are capable of a direct classification step, we will not use the classification results. \\
As the goal of the paper is to gain insight into the applicability of convolutional neural networks to this application domain, we will compare different parameter settings.\\
For the model, we restrict ourselves to YOLOv7 \citep{yolo}. While there are other types of detectors, they do not necessarily work for different datasets and are limited by the image resolution. Since our images have an image size of up to 50,000 pixels, we need detectors that scale well to higher resolutions.
YOLO-type (one-stage detectors) are widely used in real-world applications \citep{kagglechest, kagglecovid, kagglenfl, kagglecancer} and outperform other approaches such as keypoint detection \citep{DBLP:journals/corr/abs-1808-01244, DBLP:journals/corr/abs-1904-08189, DBLP:journals/corr/abs-1904-07850}. They also tend to have a faster convergence than DETR \citep{zhang2022accelerating} and work for non-photography images.\\
Apart from detectors, segmentation networks like U-Net \citep{DBLP:journals/corr/RonnebergerFB15} are also common in the microscopy domain. However, a disadvantage is that a pixel-wise annotation is much more time-consuming. Furthermore, segmentation does not scale well to higher resolutions because the input is as big as the output. For detectors, the outputs are only coordinates and not pixels. Consider a 1000x1000 segmentation mask as an example. Storing this mask in GPU memory requires keeping $1000^2$ floating-point numbers, in addition to the lower-dimensional images generated by the feature pyramid. Consequently, the memory required for this exceeds the memory needed for storing multiple vectors with bounding box coordinates.

\begin{figure*}[t]
  \centering
  \includegraphics[scale=1.5]{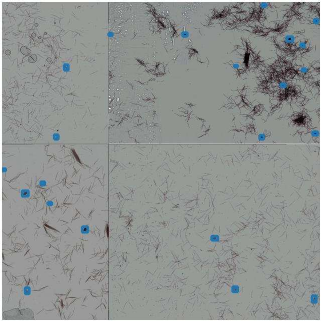}
  \caption{Mosaic data augmentation, where the blue boxes denote vessel elements}
\label{fig:mosaic}
\end{figure*}

\subsubsection{Preprocessing}

While YOLOv7 scales well to higher resolutions, there is also a limit regarding the available GPU memory. 

However, with an image of size $50000\times 50000$, vessel elements can still be identified with 10\% of the original resolution in most cases. Only to determine to which genus a vessel element belongs, one needs the full resolution to see all details. For example, \emph{Hevea} has a size of around 1400 pixels (when considering the height/width of the bounding box).

\subsubsection{Measuring the quality of object detection}
The standard metric in object detection is mean Average Precision (mAP) \citep{10.1007/s11263-009-0275-4}.

$$\text{AP} = \int_0^1 p(r) dr\,,$$

where $r$ is recall and $p(r)$ is the corresponding precision. Recall is the number of correctly found objects in relation to the total number of objects present in an image. Precision is the number of correctly found objects in relation to all found objects in an image. We only have one class that we want to detect, namely vessel elements of any type. However, when there are multiple classes, the mean over all single classes' AP is taken and called mean Average Precision (mAP).
Some additional remarks:
\begin{itemize}
    \item  we have to define the term "correctly found". For each bounding box, we compute the overlap between the prediction and the ground truth (intersection over union or short IOU). When the IOU is greater than some threshold, a found box is defined as a true positive.
    \item For $\text{AP}$, we use an IOU threshold of $0.5$.
    \item Besides mAP, we also consider precision and recall at the fixed IOU threshold $0.5$. A high recall is more important than high precision because false positives can be removed in the classification step.
    \item In practice, the integral of $\text{AP}$ is approximated by a set of eleven equally spaced recall levels \citep{10.1007/s11263-009-0275-4}. 
\end{itemize}

\subsubsection{Data augmentation}

Data augmentations are important for training YOLOv7. One example is the mosaic augmentation. Newer versions of YOLO use this augmentation because it allows to create a large amount of new training images. Three images are randomly sampled from the dataset and combined with a fourth image. Fig. \ref{fig:mosaic} illustrates how the mosaic data augmentation works. This augmentation can be used in conjunction with image or color shifts to  increase the number of possibilities for creating four images even more.

Since too many augmentations can also affect the training negatively, we restricted ourselves to a few common ones. Apart from mosaic augmentation, we used color jittering in HSV space (hue, saturation, value), image shifts, scaling and left-right flips.

\subsection{Classification}

\begin{figure*}[t]
  \centering
  \includegraphics[width=\textwidth]{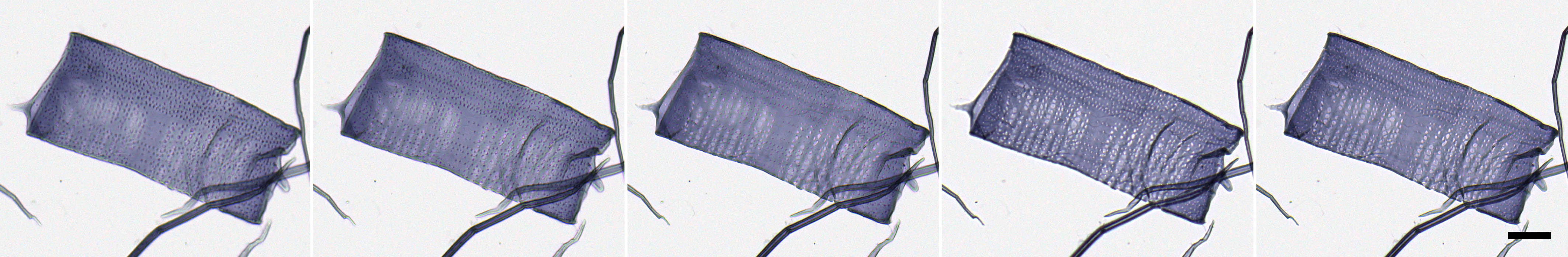}
  \caption{Comparison of focal planes for one vessel element of the genus \emph{Eucalyptus}. Depending on the focal plane, different areas (pits) of the vessel elements are in focus and therefore better visible. Scale bar=100\textmu m.}
\label{fig:focalplanes}
\end{figure*}

The task of classification is using the full resolution vessel element crop-outs from the detection step and classifying them according to their wood genus. \Cref{fig:focalplanes} shows an example of a vessel element for five different focal planes. Here, the structural differences of the individual planes, such as the vessel-ray pits or the intervessel pits, are visible. The output for each vessel element candidate is the confidence for it to belong to a specific class. For the prediction, we chose the class with the highest likelihood.

Different \textbf{architectures} are evaluated to see how the architecture influences the result. Architectures tested are ConvNeXt \citep{DBLP:journals/corr/abs-2201-03545}, EfficientNet \citep{DBLP:journals/corr/abs-1905-11946}, ResNet \citep{DBLP:journals/corr/HeZRS15} and DenseNet \citep{DBLP:journals/corr/HuangLW16a}. The dataset ImageNet is usually used for evaluating the accuracy of classification architectures. While there are many more architectures, it was shown by \citep{fang2023does} that an increase of accuracy on datasets such as ImageNet does not necessarily translate to an improvement on real-world datasets. The chosen architectures are state-of-the-art approaches that were developed in recent years.

\subsubsection{Preprocessing}

One important contribution of this paper is to provide quantitative information on how to adapt/preprocess microscopic images for hardwood identification in a way that deep neural networks can successfully be applied. We therefore identified a number of image parameters that will be analyzed with respect to their influence on performance and accuracy. \\

Above all, \textbf{resolution} is a key feature. However, similar to object detection, hardware limitations are an important issue. Higher resolution means that a smaller batch size must be used. However, a small batch size may inhibit the convergence of a neural network.
While biologically the highest image size would be best, we need to test if this is also true for neural networks.\\

Another parameter is how to \textbf{handle the variance in the size} of the detected vessel elements, i.e. the variance in image size. The neural networks expect the same image size for each vessel element. For datasets with natural images (dogs, cats, etc.), the solution is to resize the images, with the most common  size being 224x224. The reason for this is that the classes can be distinguished even if the objects in question are distorted.
However, in the case of wood identification, this could destroy important features for genus determination. An alternative approach is therefore to pad the image with zeros. The disadvantage of padding is that many pixels of the image do not provide any information. For a big vessel element, most of the image is the vessel element itself. But for a small vessel element, the image is mostly black.\\

As the vessel elements are three-dimensional objects, another preprocessing parameter is the \textbf{number of focal plane used} for classification. The microscope used to generate the images provides multiple focal planes that allow us to see specific areas of a region in more detail. To simulate the same behavior with a neural network, we try to use the focal planes as channels. Alternatively, we can input all the focus planes as individual images and combine the results.\\

Finally, we also have to consider the \textbf{importance of color} of the image. The samples were prepared with the Alexander Herzberg and nigrosin solution. These solutions affect the color of the vessel elements. It is possible that the neural networks are too biased towards the image color, while in reality it is not an important feature. Therefore, we also try grayscale images as input.

\subsubsection{Measuring the quality of vessel element classification}

The usual metric for evaluating classification is accuracy. However, this metric is biased towards the classes with the most samples. Therefore, it would be a problem here:\\
\emph{Hevea} usually has only a few vessel elements per image, whereas \emph{Populus} and \emph{Fagus} can have hundreds of vessel elements per image. Maximizing accuracy would mean ignoring \emph{Hevea}, since it does not significantly affect the objective function.\\
One solution to deal with class imbalance is to use macro F1 (averaged F1) because it highlights the performance of rare labels \citep{Lipton2014, DBLP:journals/corr/abs-1911-03347}. It is defined as

$$\text{Macro F1} = \frac{1}{n}\sum_{i=1}^n\frac{2 \cdot \text{precision}_i \cdot \text{recall}_i}{\text{precision}_i + \text{recall}_i}\,,$$

where $n$ is the number of genera. Similar to accuracy, it summarizes how the classifier performs overall. In the equation, precision and recall have the same weight. F2 is often used to give more weight to recall.

In addition to that, we will also use the so-called confusion matrix. 
In a confusion matrix, then the results are reported in more detail. Especially in a classification task, this matrix shows not only true positives (in the diagonal), but also which classes are typically confused with which other classes. 

\subsubsection{Data augmentation}

As with object detection, data augmentation is also important for classification. One can generate more data and influence the variance of the data. One typical example in case of vessel element classification is to rotate the available instances, as they can appear in arbitrary orientation.

It is important, however, to apply so-called class-preserving augmentations: some augmentations might actually change the instance in a way that leads to it no longer being representative for the respective class. For simplicity, we apply the following data augmentations as baseline: vertical flips, brightness, contrast, saturation and hue.

None of these augmentations destroy important biological features such as vessel element size. While other augmentations such as Gaussian noise or horizontal flips would also be class-preserving, this does not mean that more regularization would always lead to better results.

An example is ImageNet, architectures trained on this dataset apply only horizontal flips and no vertical flips. Therefore, we need to test which additional augmentations really improve the results.

It is not enough for an augmentation to be only class-preserving. If the neural network is able to learn a particular relationship based on the data alone, then the augmentation is unnecessary and may actually degrade the performance of the classifier. In addition, there may be a distributional shift since the real data may not contain vessel elements with e.g. Gaussian noise.

\section{Results}

\subsection{The dataset, annotation speedup}

We chose the dataset in such a way that it is sufficiently generalizable on the one hand and on the other hand defines a tackable sub-problem.

We decided to use samples from the following hardwood genera: \emph{Salix, Populus, Hevea, Fagus, Eucalyptus, Betula, Acacia, Liquidambar} and \emph{Schima}. These genera are cultivated or processed worldwide for pulp production and are commonly identified in fiber products or in case of \emph{Salix} a look-alike. With these genera, we produced an image dataset for fiber references for the first time.

\begin{figure*}[t]
\begin{subfigure}{.5\textwidth}
  \centering
  \includegraphics[scale=0.6]{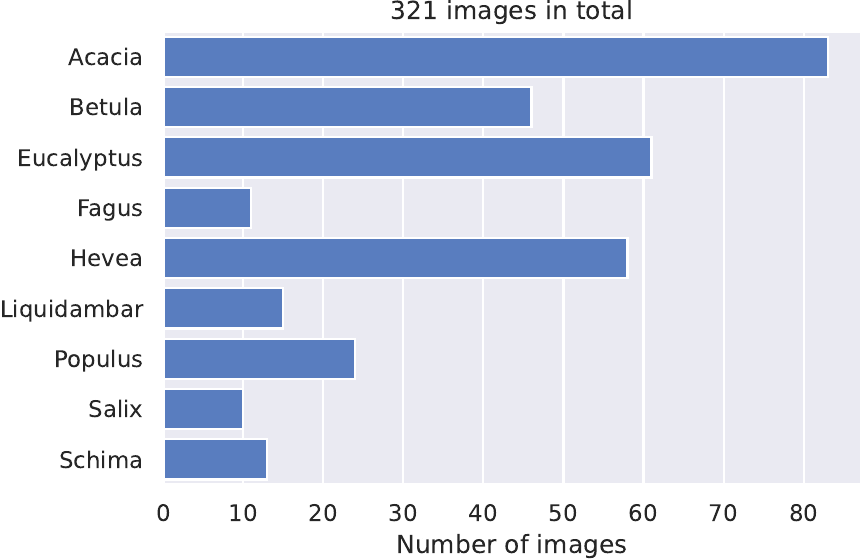}
  \label{fig:sub-first1}
\end{subfigure}
\begin{subfigure}{.5\textwidth}
  \centering
  \includegraphics[scale=0.6]{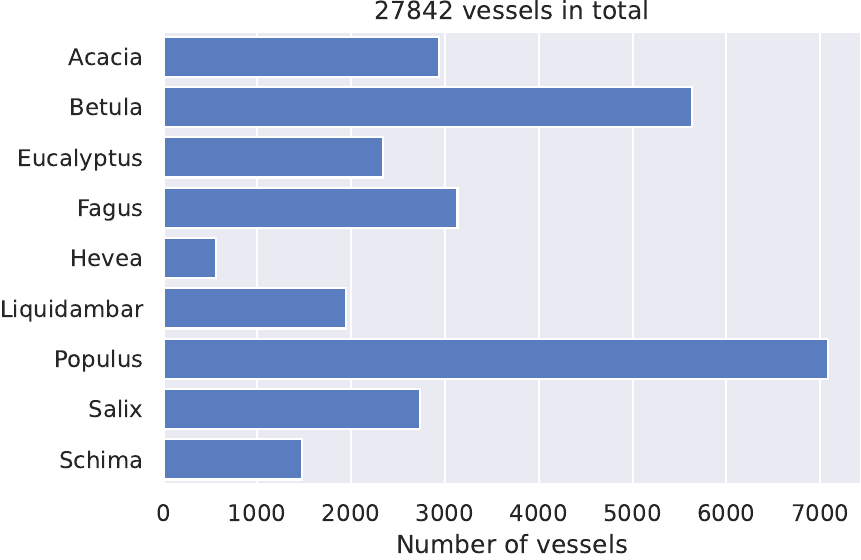} 
  \label{fig:sub-second2}
\end{subfigure}
\caption{Number of annotated images and vessels, ordered by genus}
\label{fig:DatasetStats}
\end{figure*}

\Cref{fig:DatasetStats} shows the number of images and vessel elements of the dataset. Predictions were generated with the Axioscan 7 microscope. The predictions were corrected, a new model was trained, and new predictions were made using the updated model. After repeating this procedure a couple of times, we obtained a relatively large dataset.\\
This iterative procedure facilitates annotation as described above. For each new species, some initial images were annotated without any prior prediction to make sure the detection algorithm is adapted to the specific characteristics of each species. Further annotation was done by using predicted vessel elements and only checking them. The annotation of the predicted images takes significantly less time than the full manual annotation takes.

One image including the five focal planes has a size of approx. 3.8 GiB. For this dataset, we have therefore around 1.2 TB of images. In order to train neural networks on this large amount of data, the images are preprocessed in different ways for both detection and classification, as was described before.

\newpage

\subsection{Detection results}

The smallest YOLOv7 architecture W6 has 70.4 million parameters, while the biggest one E6E has 151.7 million. We did not find that bigger models improved the detection results. Instead, it made the training more unstable. It also makes it harder to train and run the models on GPUs with small memory.

\begin{figure}[h]
  \centering
  \includegraphics[scale=0.65]{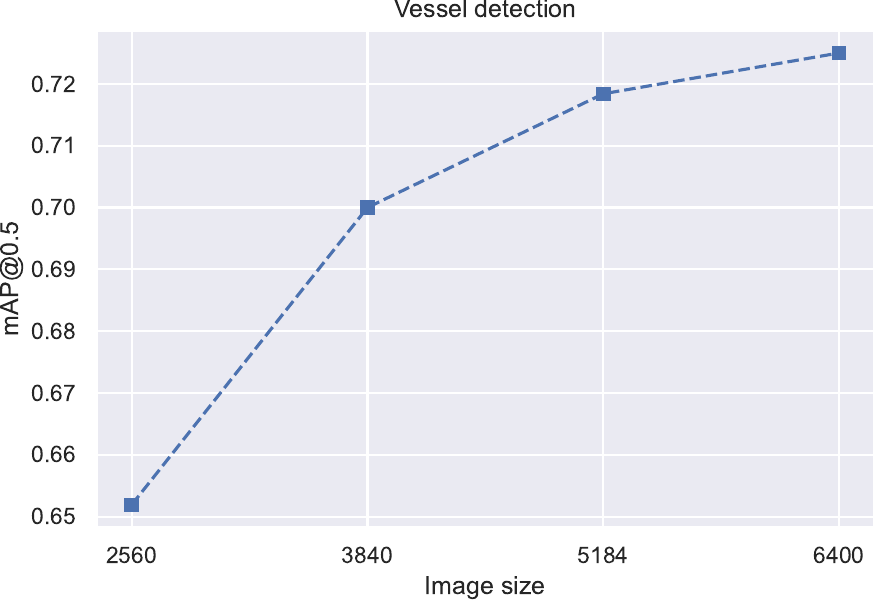}
  \caption{Effect of image size on mAP on the validation dataset}
\label{fig:imagesizedetection}
\end{figure}

A more important parameter is image size, as seen in fig. \ref{fig:imagesizedetection}. Increasing the image size up from 2560 to 6400 leads to 7\% higher mAP. This is a consequence of the model finding more vessel elements (higher recall).

After 5184 pixels, increasing the image size resulted in only minor improvements but at a considerable computational cost. Both training and prediction speeds slow down with increasing image size.

Other hyperparameters such as learning rate, number of epochs, gradient accumulation and more epochs only resulted in minor decreases or increases of mAP ($<0.5\%$).

After having determined the optimal hyperparameters, the trained model is tested on the test dataset. The results remain stable between the validation and test dataset. On the test dataset, we achieve an mAP of 71.85\% with 77.63\% precision and 72.98\% recall.

\begin{table}[ht]
        \centering
        \begin{tabular}{llll}
        \toprule
        Genus & Precision & Recall & F2 \\ \hline
        Liquidambar & 0.8885 & 0.6145 & 0.6549\\
        Salix & 0.9109 & 0.6317 & 0.6730\\
        Fagus & 0.9357 & 0.6799 & 0.7192\\
        Populus & 0.9578 & 0.6855 & 0.7268\\
        Eucalyptus & 0.8125 & 0.7629 & 0.7723\\
        Hevea & 0.5060 & 0.9037 & 0.7809\\
        Schima & 0.8736 & 0.8537 & 0.8576\\
        Betula & 0.8961 & 0.8581 & 0.8654\\
        Acacia & 0.8753 & 0.8950 & 0.8910\\
        \bottomrule
        \end{tabular}
        \caption{Detection results for individual genera, ordered by F2}
        \label{tab:detectionbyclass}
    \end{table}

Some genera produce more errors than others. This can be seen in \cref{tab:detectionbyclass}. Notably, \emph{Hevea} tends to have low precision and high recall. With \emph{Liquidambar} and \emph{Salix}, it is the other way around.

Low precision can be improved by increasing the confidence threshold. Similarly, low recall can be improved by decreasing the confidence threshold.

\begin{figure*}
    \centering
    \includegraphics[width=\textwidth]{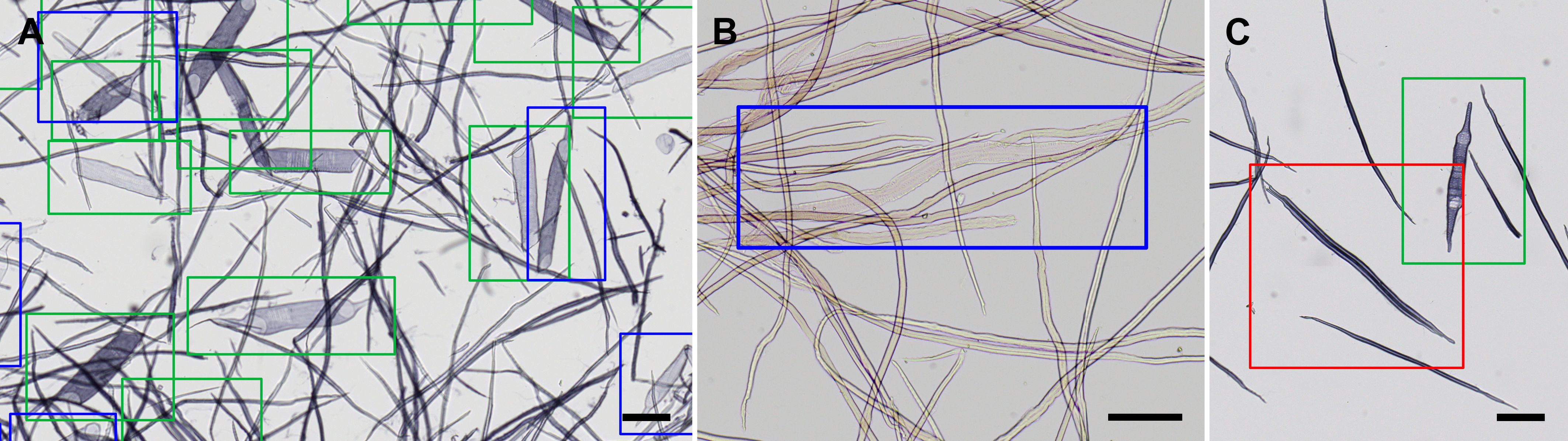}
    \caption{Typical errors. A (\emph{Fagus}): A large number of vessel elements leads to true positive (green), but also to false negative (blue). B (\emph{Liquidambar}): Low brightness or contrast causes false negative (blue), too. C (\emph{Eucalyptus}): Cohesive fibers have some similarity with vessel elements and lead to false positive (red). Scale bars=200µm}
    \label{fig:mistakes}
\end{figure*}

We mainly find three types of errors, as seen in \cref{fig:mistakes}. Some genera such as \emph{Fagus} or \emph{Liquidambar} have many vessel elements. The detector is not always able to find the vessel elements when they are too close to each other (a). Additionally, some images have low brightness (b). This leads to a low recall for certain images. Finally, there are also false positives because fibers are similar to vessel elements (c).

Real-world wood identification does not rely on single vessel element classification, as the identifiability is not always given. Therefore, to mimic real-world performance, we only need to ensure that a sufficiently high number of vessel elements is correctly detected. Then the predicted vessel  element distribution is approximately correct. In other words, it is more important to know that certain genera are in the image. It does not matter if some vessel elements are not found correctly or even classified incorrectly. Only overall, the result has to be correct.

With a larger dataset, we also expect fewer errors to occur. While we have almost 30,000 images for classification, we have 100 times fewer for detection.

\subsection{Classification results}

For all tests, we used the Adam optimizer \citep{kingma2014adam} with a batch size of 32 and a learning rate of either $10^{-3}$ or $10^{-5}$.

\begin{figure}[ht]
  \centering
  \includegraphics[scale=0.65]{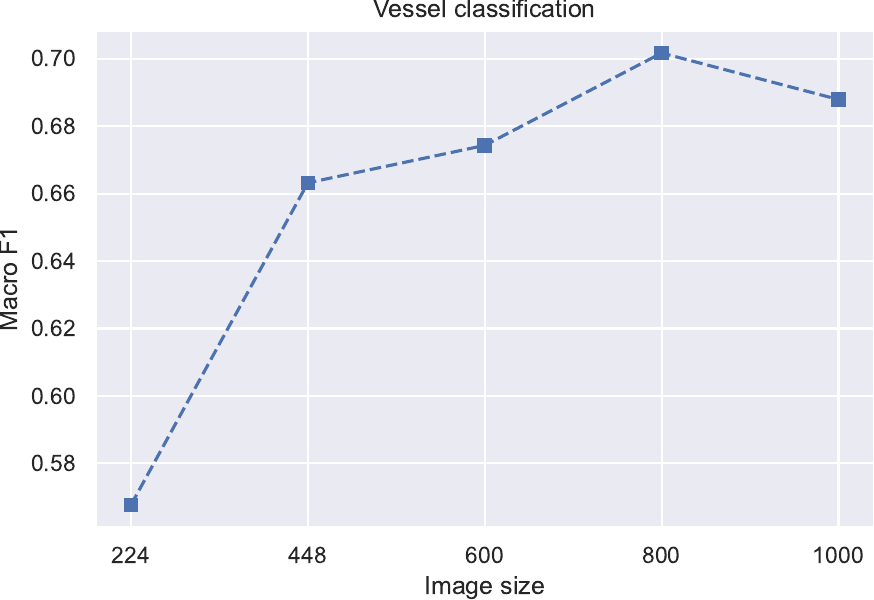}
  \caption{Effect of image size on F1 on the validation dataset}
\label{fig:imagesizeclassification}
\end{figure}

First, we determined the required image \textbf{resolution}. As \cref{fig:imagesizeclassification} shows, increasing the size from 224x224 to 800x800 improves the macro F1 score by about 12\%.

If the target image size is $800 \times 800$ pixels, we resize all images above that number so that both sides are $\leq$ 800 pixels. For example, with $1241\times 766$ pixels, we would have a new image of size $1241 \cdot r = 800$ and $800 \cdot r \approx 516$ for $r = \min\left(\frac{800}{1241}, \frac{800}{766}\right)$. The remaining $800 - 516$ pixels are padded from both sides with zeros. When for a vessel element both sides are $\leq 800$, no resizing is required, and we only pad the vessel element with zeros.

From the plot, we can see that higher sizes than $800\times 800$ did not lead to better scores. A possible explanation is that the vessel element size of many genera is on average lower than 1000 pixels. This means that many vessel elements that have a lower resolution would be padded by black pixels that do not provide any information.
\\
We tested different strategies, how to handle this \textbf{difference in image size}. One possibility is to replace the prior padding strategy by resizing. Following the previous example, the image would be resized from $800 \times 516$ to $800 \times 800$ instead of being padded. However, this leads to a distortion of the features. We find that this strategy actually decreases the macro F1 score by 1.2\%.
\\
After image size, we tested how \textbf{coloring} with the Alexander Herzberg and nigrosin solutions affected the classification. We found that converting the images to grayscale improved the score by 1\%. From a biological point of view, this result is reasonable because staining is not an important feature. From a computational point of view, it means that we can save disk space by having only one channel.
\\
Another advantage is that for the \textbf{focal plane} tests, we can work with $5$ instead of $5 \cdot 3$ channels, where $5$ is the number of focal planes.
In \cref{tab:focalplane}, we tested multiple configurations. As can be seen, there is only a marginal difference between using $1$ or $3$ channels. Using the first, second and third plane is almost as good as using only the third one.

    \begin{table}[ht]
        \centering
        \begin{tabular}{lllll}
        \toprule
        Focal plane & Macro F1 \\ \hline
1st, 2nd, 3rd & 0.6669\\
1st, 3rd, 4th & 0.6615\\
3rd & 0.6602\\
Average & \textbf{0.7017}\\
Maximum & 0.7014\\
        \bottomrule
        \end{tabular}
        \caption{Focal plane tests}
        \label{tab:focalplane}
        \end{table}

Therefore, the better approach is to consider only a single channel and do multiple predictions. In the table, there are two combination strategies for the probabilities. Either averaging the five probability vectors, or taking element-wise the maximum of the values. Both approaches perform similarly.

It can be argued that the focal planes represent a kind of test-time augmentation (TTA). Some regions are sharpened, but the overall image distribution remains the same. TTA is often used in microscopy and other fields to improve results \citep{Moshkov2020}.

After general tests with respect to preprocessing, we tried various architectures to see how it affects the macro f1 score.
        
    \begin{table}[ht]
        \centering
        \begin{tabular}{lllll}
        \toprule
        Architecture & Macro F1 \\ \hline
ConvNeXt-tiny & \textbf{0.7017}\\
DenseNet-121 & 0.6441\\
ResNet-34 & 0.5958\\
EfficientNet-B0 & 0.6472\\
EfficientNet-B1 & 0.6698\\
EfficientNet-B2 & 0.6632\\
        \bottomrule
        \end{tabular}
        \caption{Architecture experiments}
        \label{tab:architecture}
        \end{table}

As can be seen from \cref{tab:architecture}, ConvNeXt-tiny leads to the best results. This architecture also performed best on other real-world datasets, as was shown by \cite{fang2023does}. We found that ConvNeXt-tiny has problems converging, with a learning rate of $10^{-3}$. Therefore, we used for this architecture $10^{-5}$.

Finally, we tested adding more class-preserving data augmentations like horizontal flips or Gaussian noise. Applying two types of flips reduces macro F1 by about 1\%. This means that if the model is already capable of learning the rotation based on the data alone, then adding more data augmentation may degrade performance. Gaussian noise also did not show any increase of macro F1.

After having determined the best hyperparameters, we run the models on the test dataset. We achieve a macro F1 score of 64.61\% on the test dataset. It is slightly lower than that of the validation dataset. The reason is that the vessel element distribution is not exactly the same as the one of the validation dataset. Some classes are underrepresented.

\begin{figure}[ht]
  \centering
  \includegraphics[scale=0.75]{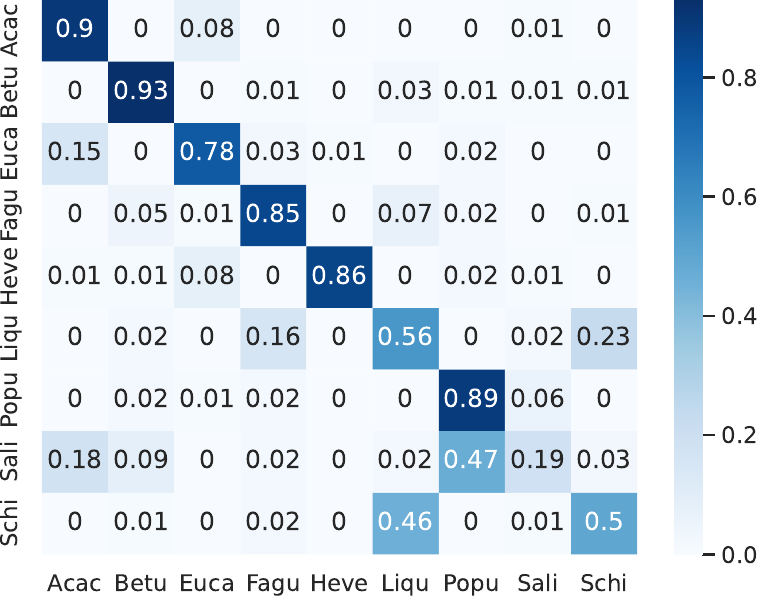}
  \caption{Confusion matrix}
\label{fig:confusionmatrix}
\end{figure}

Finally, we look at the confusion matrix to see the performance across different classes. \Cref{fig:confusionmatrix} shows that the following genera are often confused: \emph{Liquidambar--Schima--Fagus} and \emph{Populus-Salix}. These genera are characterized by great similarities, and not every vessel element shows all structural features. Thus, even for wood anatomists, it is difficult to make a clear classification for each individual cell. Therefore, our model behaves similarly to a human expert. 

When only considering the diagonal, we see that most genera are above 80\%. The only problematic genera are \emph{Liquidambar, Salix} and \emph{Schima}. But as these classes are even difficult to classify for each vessel element by humans, this confirms that our model is working as intended. If it is not possible to classify these genera even as a human, then the genera can also be combined into one class. This reduces the number of classes and increases the F1 score. In terms of accuracy, it is therefore possible to obtain a classifier with an accuracy of over 90\%.


\section{Discussion}

\subsection{Pipeline}

In general, the pipeline is well adapted to be used on such a problem. We have been able to generate a for now big data basis for further analysis. We can use a plethora of different detection and classification methods -- e.g. various neural networks -- and directly compare their results.

It might be argued that by mimicking the visual manual process, we limit the capacity of the solution to detect species by using other cell structures than vessel elements. However, the database as we build it, could also be used in the future to be combined with mixed overview images, as the vessel element detector works on different vessel element types. Furthermore, we now have a first version of a mixed species image detector/classifier that could again be used in a later stage for assisted annotation of mixed samples.

\subsection{Detection}

For detection, we compared how image size and larger architectures affect detection results. It is better to use higher resolutions but smaller architectures.

Although large models like YOLOv7-E6E are available, they do not improve the mean average precision (mAP) of identifying vessel elements as this task does not require complicated features. The depth of a neural network, which is associated with the receptive field, can impact its ability to "see" more of an image 
\citep{Araujo2019, DBLP:journals/corr/LuoLUZ17}. A higher receptive field can be achieved by stacking more convolutions, which leads to a higher effective kernel size and allows for the processing of more pixels at once. This can be useful when detecting small details.

However, in this case, using a large model was not necessary as the vessel element identification task does not require big complicated features. Therefore, a simpler model is sufficient for accurately identifying vessel elements, and using a larger model would not provide any significant benefits.

The data itself has a big effect on what the neural network is able to "see". If the vessel elements are difficult to see with the human eye (e.g. due to low brightness), the detection performance of the detector also decreases significantly. Therefore, it is important to make sure that manual preparation process as well as the images produced by the microscope are of good quality. In contrast to recall ("how many objects are found?"), precision ("is the object found really an object?") is usually good, regardless of how dark or light the vessel elements are.

Furthermore, even if an area is misidentified as a vessel element, it is possible that the classifier can still determine the genus. For now, we have only modeled our pipeline based on current biological domain knowledge. Other unknown features could also indicate the genus of a vessel element.

Therefore, it is more important to have a high recall. The recall rates of certain genera are quite low at 60\%. However, one must always keep in mind that hundreds of vessel elements have already been found correctly for these genera. Only when an image contains few vessel elements and the detector misses them, a low recall is problematic.

While annotating the data, we found that having multiple vessel elements on top of each other was a problem. It would be necessary for the bounding boxes to be able to be rotated as well. Otherwise, we would have only one bounding box covering both bounding boxes. The other option is a segmentation mask or vertex-based detection. However, this problem only affects genera like \emph{Liquidambar} where we already have a large amount of vessel elements.

In general, the object detector already provides sufficiently good results. We expect recall and precision to improve with more labeled data.

\subsection{Classification}

For classification, we found that data leakage was initially a major problem. Instead of focusing on vessel elements, the neural networks paid attention to brightness. By carefully splitting the data based on maceration ID and maintaining the same data distribution between the training and validation datasets, we were able to solve this problem. In addition, as the size of the dataset increased, the generalization performance of the classifier also improved. Data augmentations such as brightness or contrast also prevents the network on just focusing on the background.

Since, from a biological perspective, the information leading to a particular classification is clear, one could also consider modeling the methods to focus on specific areas of the vessel elements (such as the vessel-ray pits). However, as already previously mentioned, it is possible to find additional features that could indicate the genus.

The current worst performing genera are also those that are confused by human experts. Therefore, the classifier produces the expected results. We did, however, not surpass human performance on vessel element classification based on our preliminary tests.

Our extensive hyperparameter tuning was guided by biological domain expertise. We found that focal planes and high image resolution were important for the classification. Converting the images from RGB to grayscale also improved macro F1.

Apart from the preprocessing, the chosen architecture also makes a difference. ConvNeXt produced the best results, while smaller architectures such as EfficientNet-B0 performed slightly worse. This is unlike detection, where the smaller models tended to work better.

Since we use a high resolution of 800 pixels, the architecture also has to be deeper to be able to see all the important features. We need therefore a higher receptive field such that the network is able to learn features which are also important for humans. Experts also require high-resolution images to have a sufficient level of detail to find the relevant features for distinguishing genera. For object detection, shallower networks worked better because a high level of detail is not required to determine if an area contains a vessel element.

\section{Conclusions}

We have shown that wood detection in microscopy images can be automatized with neural networks. We have performed extensive evaluation of hyperparameters to ensure the representativity and robustness of our results. Our method achieved similar results to human experts. Extensive tests have shown that the genera that are usually confused by humans are also problematic for neural networks. The object detection sometimes misses vessel elements, but the performance is already good enough to produce a large candidate list for the classifier.

In future work, we want to see whether the performance can be extended to more genera. 

Additionally, we do not know yet which regions the network really focuses on. A more in-depth analysis of the neural networks areas of focus on these microscopic images would be of interest.

It might be possible to discover new features important for the classification by looking at class activation maps or saliency maps \citep{DBLP:journals/corr/SundararajanTY17}.

Finally, we will produce and evaluate  mixed samples containing multiple genera. We also intend to perform a blind test, AI versus anatomists.

\section{Competing interests}
No competing interest is declared.

\section{Author contributions statement}
L.N., H.S. conceived the experiment(s),  L.N. conducted the experiment(s), A.O., J.S.-R. and S.H. conceived data generation of references. All authors analyzed the results. All authors wrote and reviewed the manuscript.

\section{Acknowledgments}
The authors would like to thank all colleagues who participated in the preparation of the numerous samples, helped with annotation and made the project happen: P.Gospodnetić, L.Gradert, J.Heddier, D.Helm, S.Kaschuro, G.Koch, C.Piehl, M.Rauhut, L.Wenrich, A. Wettich, S.Wrage (all Fraunhofer Institute for Industrial Mathematics ITWM or Thünen Institute of Wood Research). This work is supported by funds from the Fachagentur Nachwachsende Rohstoffe e.V. (FNR - FKZ 2220HV063A and 2220HV063B)

\bibliographystyle{unsrtnat}
\bibliography{references}  

\begin{thebibliography}{44}
\providecommand{\natexlab}[1]{#1}
\providecommand{\url}[1]{\texttt{#1}}
\expandafter\ifx\csname urlstyle\endcsname\relax
  \providecommand{\doi}[1]{doi: #1}\else
  \providecommand{\doi}{doi: \begingroup \urlstyle{rm}\Url}\fi

\bibitem[{European Commission}(2021)]{european2021proposal}
{European Commission}.
\newblock Proposal for a regulation of the european parliament and of the
  council on the making available on the union market as well as export from
  the union of certain commodities and products associated with deforestation
  and forest degradation and repealing regulation (eu) no 995/2010.
\newblock 2021.

\bibitem[Schmitz et~al.(2020)Schmitz, Beeckman, Blanc-Jolivet, Boeschoten,
  Braga, Cabezas, Chaix, Crameri, Deklerck, Degen, et~al.]{schmitz2020overview}
Nele Schmitz, Hans Beeckman, Celine Blanc-Jolivet, Laura Boeschoten, Jez~WB
  Braga, Jos{\'e}~Antonio Cabezas, Gilles Chaix, Simon Crameri, Victor
  Deklerck, Bernd Degen, et~al.
\newblock Overview of current practices in data analysis for wood
  identification. a guide for the different timber tracking methods.
\newblock Technical report, 2020.

\bibitem[Wheeler et~al.(1989)Wheeler, Baas, Gasson, et~al.]{wheeler1989iawa}
Elisabeth~A Wheeler, Pieter Baas, Peter~E Gasson, et~al.
\newblock Iawa list of microscopic features for hardwood identification.
\newblock Technical report, 1989.

\bibitem[Richter and Dallwitz(2000-onwards)]{Jorgo}
Hans~Georg Richter and Dallwitz.
\newblock Commercial timbers: Descriptions, illustrations, identification, and
  information retrieval, 2000-onwards.
\newblock URL \url{https://www.delta-intkey.com/wood/en/index.htm}.
\newblock (accessed on 15 May 2023).

\bibitem[Wheeler(2004-onwards)]{insidewood}
Elisabeth~A Wheeler.
\newblock Insidewood, 2004-onwards.
\newblock URL
  \url{https://insidewood.lib.ncsu.edu/search;jsessionid=QlhMElom39gUT2qFJDnuE6vLudO1GnIyCEP340Bx?0}.
\newblock (accessed on 15 May 2023).

\bibitem[Koch and Koch(2022)]{openagrar_mods_00085331}
Gerald Koch and Sven Koch.
\newblock Holzartenwissen im app-format : neue app "macroholzdata" zur
  holzartenbestimmung und -beschreibung.
\newblock \emph{Furnier-Magazin}, 26:\penalty0 52--56, 2022.
\newblock URL \url{https://www.openagrar.de/receive/openagrar_mods_00085331}.

\bibitem[Ilvessalo-Pf{\"a}ffli(1995)]{ilvessalo1995fiber}
Marja-Sisko Ilvessalo-Pf{\"a}ffli.
\newblock \emph{Fiber atlas: identification of papermaking fibers}.
\newblock Springer Science \& Business Media, 1995.

\bibitem[Helmling et~al.(2018)Helmling, Olbrich, Heinz, and
  Koch]{helmling2018atlas}
Stephanie Helmling, Andrea Olbrich, Immo Heinz, and Gerald Koch.
\newblock Atlas of vessel elements: Identification of asian timbers.
\newblock \emph{Iawa Journal}, 39\penalty0 (3):\penalty0 249--352, 2018.

\bibitem[Ruffinatto and Crivellaro(2019)]{ruffinatto2019atlas}
Flavio Ruffinatto and Alan Crivellaro.
\newblock \emph{Atlas of macroscopic wood identification: with a special focus
  on timbers used in Europe and CITES-listed species}.
\newblock Springer Nature, 2019.

\bibitem[Richter et~al.(2014-onwards)Richter, Gembruch, and Koch]{JorgoCITES}
Hans~Georg Richter, Karin Gembruch, and Gerald Koch.
\newblock Citeswoodid: descriptions, illustrations, identification, and
  information retrieval, 2014-onwards.
\newblock URL \url{https://www.delta-intkey.com/citeswood/en/index.htm}.
\newblock (accessed on 15 May 2023).

\bibitem[UTAR and FRIM(2018)]{MyWood}
UTAR and FRIM.
\newblock Mywood-premium, 2018.
\newblock URL \url{https://mywoodid.frim.gov.my/}.
\newblock (accessed on 15 May 2023).

\bibitem[Ravindran et~al.(2020)Ravindran, Thompson, Soares, and
  Wiedenhoeft]{ravindran2020xylotron}
Prabu Ravindran, Blaise~J Thompson, Richard~K Soares, and Alex~C Wiedenhoeft.
\newblock The xylotron: flexible, open-source, image-based macroscopic field
  identification of wood products.
\newblock \emph{Frontiers in plant science}, 11:\penalty0 1015, 2020.

\bibitem[Wiedenhoeft(2020)]{wiedenhoeft2020xylophone}
Alex~C Wiedenhoeft.
\newblock The xylophone: toward democratizing access to high-quality
  macroscopic imaging for wood and other substrates.
\newblock \emph{Iawa Journal}, 41\penalty0 (4):\penalty0 699--719, 2020.

\bibitem[Silva et~al.(2022)Silva, Bordalo, Pissarra, and
  de~Palacios]{silva2022computer}
Jos{\'e}~Lu{\'\i}s Silva, Rui Bordalo, Jos{\'e} Pissarra, and Paloma
  de~Palacios.
\newblock Computer vision-based wood identification: A review.
\newblock \emph{Forests}, 13\penalty0 (12):\penalty0 2041, 2022.

\bibitem[Franklin(1945)]{franklin1945preparation}
GL~Franklin.
\newblock Preparation of thin sections of synthetic resins and wood-resin
  composites, and a new macerating method for wood.
\newblock \emph{Nature}, 155\penalty0 (3924):\penalty0 51--51, 1945.

\bibitem[Helmling et~al.(2016)Helmling, Olbrich, Tepe, and
  Koch]{helmling2016qualitative}
Stephanie Helmling, Andrea Olbrich, Lena Tepe, and Gerald Koch.
\newblock Qualitative and quantitative characteristics of macerated vessels of
  23 mixed tropical hardwood {(MTH)} species: a data collection for the
  identification of wood species in pulp and paper.
\newblock \emph{Holzforschung}, 70\penalty0 (9):\penalty0 839--844, 2016.

\bibitem[O’Mahony et~al.(2019)O’Mahony, Campbell, Carvalho, Harapanahalli,
  Hernandez, Krpalkova, Riordan, and Walsh]{o2019deep}
Niall O’Mahony, Sean Campbell, Anderson Carvalho, Suman Harapanahalli,
  Gustavo~Velasco Hernandez, Lenka Krpalkova, Daniel Riordan, and Joseph Walsh.
\newblock Deep learning vs. traditional computer vision.
\newblock In \emph{Science and information conference}, pages 128--144.
  Springer, 2019.

\bibitem[Ren et~al.(2015)Ren, He, Girshick, and
  Sun]{DBLP:journals/corr/RenHG015}
Shaoqing Ren, Kaiming He, Ross~B. Girshick, and Jian Sun.
\newblock Faster {R-CNN:} towards real-time object detection with region
  proposal networks.
\newblock \emph{CoRR}, abs/1506.01497, 2015.
\newblock URL \url{http://arxiv.org/abs/1506.01497}.

\bibitem[Caron et~al.(2021)Caron, Touvron, Misra, J{\'{e}}gou, Mairal,
  Bojanowski, and Joulin]{DBLP:journals/corr/abs-2104-14294}
Mathilde Caron, Hugo Touvron, Ishan Misra, Herv{\'{e}} J{\'{e}}gou, Julien
  Mairal, Piotr Bojanowski, and Armand Joulin.
\newblock Emerging properties in self-supervised vision transformers.
\newblock \emph{CoRR}, abs/2104.14294, 2021.
\newblock URL \url{https://arxiv.org/abs/2104.14294}.

\bibitem[Zhang et~al.(2022{\natexlab{a}})Zhang, Li, Liu, Zhang, Su, Zhu, Ni,
  and Shum]{zhang2022dino}
Hao Zhang, Feng Li, Shilong Liu, Lei Zhang, Hang Su, Jun Zhu, Lionel~M. Ni, and
  Heung-Yeung Shum.
\newblock {DINO: DETR with Improved DeNoising Anchor Boxes for End-to-End
  Object Detection}, 2022{\natexlab{a}}.

\bibitem[Wang et~al.(2022)Wang, Bochkovskiy, and Liao]{yolo}
Chien-Yao Wang, Alexey Bochkovskiy, and Hong-Yuan~Mark Liao.
\newblock Yolov7: Trainable bag-of-freebies sets new state-of-the-art for
  real-time object detectors, 2022.
\newblock URL \url{https://arxiv.org/abs/2207.02696}.

\bibitem[Hannun et~al.(2021)Hannun, Guo, and van~der
  Maaten]{DBLP:journals/corr/abs-2102-11673}
Awni~Y. Hannun, Chuan Guo, and Laurens van~der Maaten.
\newblock Measuring data leakage in machine-learning models with fisher
  information.
\newblock \emph{CoRR}, abs/2102.11673, 2021.
\newblock URL \url{https://arxiv.org/abs/2102.11673}.

\bibitem[Kaggle(2021{\natexlab{a}})]{kagglechest}
Kaggle.
\newblock {VinBigData Chest X-ray Abnormalities Detection}, 2021{\natexlab{a}}.
\newblock URL
  \url{https://www.kaggle.com/competitions/vinbigdata-chest-xray-abnormalities-detection/}.

\bibitem[Kaggle(2021{\natexlab{b}})]{kagglecovid}
Kaggle.
\newblock {SIIM-FISABIO-RSNA COVID-19 Detection}, 2021{\natexlab{b}}.
\newblock URL
  \url{https://www.kaggle.com/competitions/siim-covid19-detection/}.

\bibitem[Kaggle(2021{\natexlab{c}})]{kagglenfl}
Kaggle.
\newblock {NFL 1st and Future - Impact Detection}, 2021{\natexlab{c}}.
\newblock URL \url{https://www.kaggle.com/competitions/nfl-impact-detection/}.

\bibitem[Kaggle(2023)]{kagglecancer}
Kaggle.
\newblock {RSNA Screening Mammography Breast Cancer Detection}, 2023.
\newblock URL
  \url{https://www.kaggle.com/competitions/rsna-breast-cancer-detection/}.

\bibitem[Law and Deng(2018)]{DBLP:journals/corr/abs-1808-01244}
Hei Law and Jia Deng.
\newblock {CornerNet}: Detecting objects as paired keypoints.
\newblock \emph{CoRR}, abs/1808.01244, 2018.
\newblock URL \url{http://arxiv.org/abs/1808.01244}.

\bibitem[Duan et~al.(2019)Duan, Bai, Xie, Qi, Huang, and
  Tian]{DBLP:journals/corr/abs-1904-08189}
Kaiwen Duan, Song Bai, Lingxi Xie, Honggang Qi, Qingming Huang, and Qi~Tian.
\newblock {CenterNet}: Keypoint triplets for object detection.
\newblock \emph{CoRR}, abs/1904.08189, 2019.
\newblock URL \url{http://arxiv.org/abs/1904.08189}.

\bibitem[Zhou et~al.(2019)Zhou, Wang, and
  Kr{\"{a}}henb{\"{u}}hl]{DBLP:journals/corr/abs-1904-07850}
Xingyi Zhou, Dequan Wang, and Philipp Kr{\"{a}}henb{\"{u}}hl.
\newblock Objects as points.
\newblock \emph{CoRR}, abs/1904.07850, 2019.
\newblock URL \url{http://arxiv.org/abs/1904.07850}.

\bibitem[Zhang et~al.(2022{\natexlab{b}})Zhang, Luo, Yu, Cui, and
  Lu]{zhang2022accelerating}
Gongjie Zhang, Zhipeng Luo, Yingchen Yu, Kaiwen Cui, and Shijian Lu.
\newblock Accelerating detr convergence via semantic-aligned matching,
  2022{\natexlab{b}}.

\bibitem[Ronneberger et~al.(2015)Ronneberger, Fischer, and
  Brox]{DBLP:journals/corr/RonnebergerFB15}
Olaf Ronneberger, Philipp Fischer, and Thomas Brox.
\newblock U-net: Convolutional networks for biomedical image segmentation.
\newblock \emph{CoRR}, abs/1505.04597, 2015.
\newblock URL \url{http://arxiv.org/abs/1505.04597}.

\bibitem[Everingham et~al.(2010)Everingham, Gool, Williams, Winn, and
  Zisserman]{10.1007/s11263-009-0275-4}
Mark Everingham, Luc Gool, Christopher~K. Williams, John Winn, and Andrew
  Zisserman.
\newblock The {Pascal Visual Object Classes (VOC) Challenge}.
\newblock \emph{Int. J. Comput. Vision}, 88\penalty0 (2):\penalty0 303–338,
  jun 2010.
\newblock ISSN 0920-5691.
\newblock \doi{10.1007/s11263-009-0275-4}.
\newblock URL \url{https://doi.org/10.1007/s11263-009-0275-4}.

\bibitem[Liu et~al.(2022)Liu, Mao, Wu, Feichtenhofer, Darrell, and
  Xie]{DBLP:journals/corr/abs-2201-03545}
Zhuang Liu, Hanzi Mao, Chao{-}Yuan Wu, Christoph Feichtenhofer, Trevor Darrell,
  and Saining Xie.
\newblock A convnet for the 2020s.
\newblock \emph{CoRR}, abs/2201.03545, 2022.
\newblock URL \url{https://arxiv.org/abs/2201.03545}.

\bibitem[Tan and Le(2019)]{DBLP:journals/corr/abs-1905-11946}
Mingxing Tan and Quoc~V. Le.
\newblock {EfficientNet}: Rethinking model scaling for convolutional neural
  networks.
\newblock \emph{CoRR}, abs/1905.11946, 2019.
\newblock URL \url{http://arxiv.org/abs/1905.11946}.

\bibitem[He et~al.(2015)He, Zhang, Ren, and Sun]{DBLP:journals/corr/HeZRS15}
Kaiming He, Xiangyu Zhang, Shaoqing Ren, and Jian Sun.
\newblock Deep residual learning for image recognition.
\newblock \emph{CoRR}, abs/1512.03385, 2015.
\newblock URL \url{http://arxiv.org/abs/1512.03385}.

\bibitem[Huang et~al.(2016)Huang, Liu, and
  Weinberger]{DBLP:journals/corr/HuangLW16a}
Gao Huang, Zhuang Liu, and Kilian~Q. Weinberger.
\newblock Densely connected convolutional networks.
\newblock \emph{CoRR}, abs/1608.06993, 2016.
\newblock URL \url{http://arxiv.org/abs/1608.06993}.

\bibitem[Fang et~al.(2023)Fang, Kornblith, and Schmidt]{fang2023does}
Alex Fang, Simon Kornblith, and Ludwig Schmidt.
\newblock Does progress on {ImageNet} transfer to real-world datasets?, 2023.

\bibitem[Lipton et~al.(2014)Lipton, Elkan, and Naryanaswamy]{Lipton2014}
Zachary~C. Lipton, Charles Elkan, and Balakrishnan Naryanaswamy.
\newblock Optimal thresholding of classifiers to maximize f1 measure.
\newblock In \emph{Machine Learning and Knowledge Discovery in Databases},
  pages 225--239. Springer Berlin Heidelberg, 2014.
\newblock \doi{10.1007/978-3-662-44851-9_15}.
\newblock URL \url{https://doi.org/10.1007/978-3-662-44851-9_15}.

\bibitem[Opitz and Burst(2019)]{DBLP:journals/corr/abs-1911-03347}
Juri Opitz and Sebastian Burst.
\newblock Macro {F1} and macro {F1}.
\newblock \emph{CoRR}, abs/1911.03347, 2019.
\newblock URL \url{http://arxiv.org/abs/1911.03347}.

\bibitem[Kingma and Ba(2014)]{kingma2014adam}
Diederik~P Kingma and Jimmy Ba.
\newblock Adam: A method for stochastic optimization.
\newblock \emph{arXiv preprint arXiv:1412.6980}, 2014.

\bibitem[Moshkov et~al.(2020)Moshkov, Mathe, Kertesz-Farkas, Hollandi, and
  Horvath]{Moshkov2020}
Nikita Moshkov, Botond Mathe, Attila Kertesz-Farkas, Reka Hollandi, and Peter
  Horvath.
\newblock Test-time augmentation for deep learning-based cell segmentation on
  microscopy images.
\newblock \emph{Scientific Reports}, 10\penalty0 (1), March 2020.
\newblock \doi{10.1038/s41598-020-61808-3}.
\newblock URL \url{https://doi.org/10.1038/s41598-020-61808-3}.

\bibitem[Araujo et~al.(2019)Araujo, Norris, and Sim]{Araujo2019}
Andre Araujo, Wade Norris, and Jack Sim.
\newblock Computing receptive fields of convolutional neural networks.
\newblock \emph{Distill}, 4\penalty0 (11), November 2019.
\newblock \doi{10.23915/distill.00021}.
\newblock URL \url{https://doi.org/10.23915/distill.00021}.

\bibitem[Luo et~al.(2017)Luo, Li, Urtasun, and
  Zemel]{DBLP:journals/corr/LuoLUZ17}
Wenjie Luo, Yujia Li, Raquel Urtasun, and Richard~S. Zemel.
\newblock Understanding the effective receptive field in deep convolutional
  neural networks.
\newblock \emph{CoRR}, abs/1701.04128, 2017.
\newblock URL \url{http://arxiv.org/abs/1701.04128}.

\bibitem[Sundararajan et~al.(2017)Sundararajan, Taly, and
  Yan]{DBLP:journals/corr/SundararajanTY17}
Mukund Sundararajan, Ankur Taly, and Qiqi Yan.
\newblock Axiomatic attribution for deep networks.
\newblock \emph{CoRR}, abs/1703.01365, 2017.
\newblock URL \url{http://arxiv.org/abs/1703.01365}.

\end{thebibliography}






\end{document}